\documentclass{article}

\PassOptionsToPackage{numbers, compress}{natbib}
\usepackage[main, final]{neurips_2026}

\usepackage[utf8]{inputenc} 
\usepackage[T1]{fontenc}    
\usepackage{hyperref}       
\usepackage{url}            
\usepackage{booktabs}       
\usepackage{amsfonts}       
\usepackage{nicefrac}       
\usepackage{microtype}      
\usepackage{xcolor}         
\usepackage{graphicx}
\usepackage{amssymb}
\usepackage{algorithm}
\usepackage{algpseudocode}
\usepackage{amsmath}        
\usepackage[capitalize]{cleveref}
\usepackage{wrapfig}
\usepackage{float}
\usepackage{caption}
\usepackage{multirow}
\usepackage{booktabs}
\usepackage{enumitem}
\usepackage{graphicx}      
\usepackage{amsmath}       
\usepackage{multirow}      
\usepackage{booktabs}      
\usepackage[table]{xcolor}
\usepackage{array}
\usepackage{graphicx}
\usepackage{booktabs}
\usepackage{wrapfig}
\usepackage[table]{xcolor}
\usepackage{multirow} 
\usepackage{pifont}
\usepackage{caption}
\usepackage{float}
\usepackage{subcaption}
\title{HyperDiT: Hyper-Connected Transformers for High-Fidelity Pixel-Space Diffusion}

\author{
    \textbf{Yu He}\thanks{Equal contribution}\quad
    \textbf{Lichen Ma}\footnotemark[1]\quad
    \textbf{Zipeng Guo}\footnotemark[1]\quad
    \textbf{Xinyuan Shan}\quad 
    \textbf{Jingling Fu} \\
    \textbf{Dong Chen}\quad
    \textbf{Junshi Huang}\thanks{Corresponding Author.}\quad
    \textbf{Yan Li} \\
    {\tt\small \{heyu2579, junshi.huang\}@gmail.com}
}

\begin{document}

\maketitle

\begin{abstract}
    Pixel-space diffusion models bypass the reconstruction bottleneck of Variational Autoencoders (VAEs) but face a fundamental "granularity dilemma": capturing global semantics favors large patch scales, while generating high-fidelity details demands fine-grained inputs. 
    To address this issue, we propose \textbf{HyperDiT}, a unified framework establishing \emph{Hyper-Connected Cross-Scale Interactions} to bridge the semantic and pixel manifold. 
    Diverging from injecting semantics by AdaLN, HyperDiT utilizes Cross-Attention mechanisms, enabling fine-grained tokens to query multi-level semantic anchors globally. 
    To resolve the spatial mismatch during multi-scale interactions, we introduce Scale-Aware Rotary Position Embedding (SA-RoPE) to ensure precise geometric alignment among tokens of varying patch sizes. 
    Furthermore, we incorporate Registers to learn the dense semantics from a pretrained Visual Foundation Model (VFM), effectively reducing generation hallucination and artifacts. 
    Extensive experiments demonstrate that HyperDiT achieves state-of-the-art (SoTA) FID of $\mathbf{1.56}$ on ImageNet $256\times256$ directly within the pixel space. 
    By combining the fine-grained stream with semantic guidance, HyperDiT offers a superior paradigm for high-fidelity pixel generation.
\end{abstract}

\section{Introduction}
\label{sec:intro}
Diffusion models \cite{ho2020denoising, song2020denoising} have recently driven remarkable progress in image generation. 
To manage the computational overhead of generating high-resolution images, SoTA approaches typically adopt the latent diffusion paradigm \cite{rombach2022high,ma2024sit,zheng2025diffusion}, relying on VAEs \cite{kingma2013auto} to generate images in a low-dimensional latent space. 
However, VAEs introduce an inherent reconstruction bottleneck, leading to irreversible loss of high-frequency details and structural artifacts that limit the upper bound of image fidelity \cite{podell2023sdxl}. 
Recently, works such as JiT \cite{jit} and DeCo \cite{ma2025deco} have advocated for returning to the native pixel space. 
By discarding the VAE and modeling the natural data distribution directly, pixel-space diffusion models unlock the potential for extreme generation fidelity.

\begin{wrapfigure}[20]{r}{0.4\textwidth}
    \centering
    \includegraphics[width=0.4\textwidth]{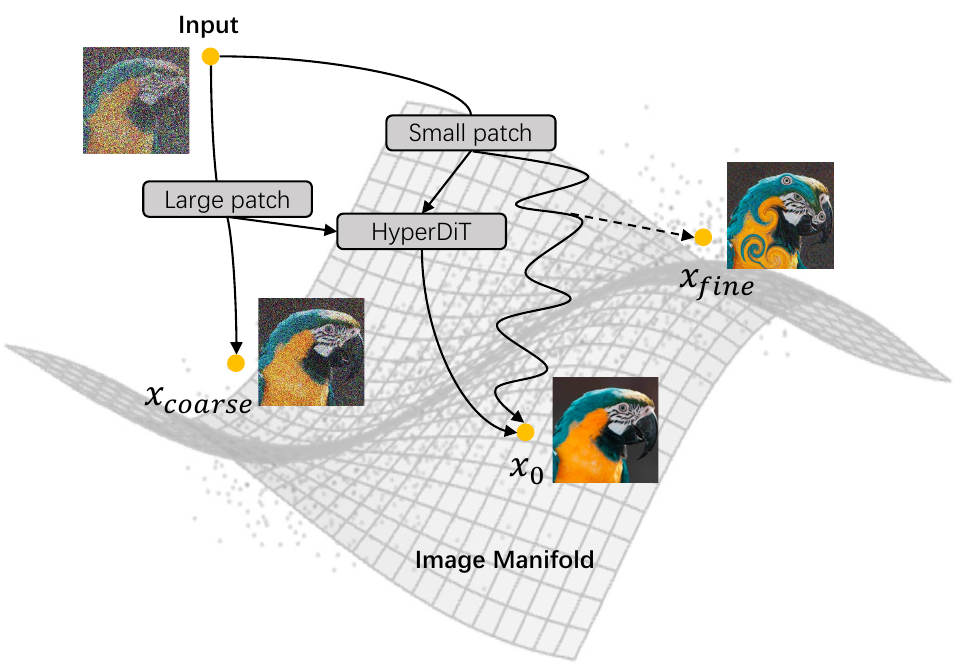}
    \caption{Conceptual illustration of generation trajectories. Large patches ($x_{coarse}$) fail to capture fine details, whereas small patches ($x_{fine}$) struggle with global coherence. Our proposed HyperDiT leverages dense cross-scale interactions to guide the generation process, landing on the image manifold ($x_{0}$).}
    \label{fig:manifold}
\end{wrapfigure}

However, generating images directly in pixel space presents a severe "granularity dilemma". 
From a manifold learning perspective, the generation process is essentially guiding random noise along a trajectory that converges to the natural image manifold. 
As illustrated in \cref{fig:manifold}, large patches provide a macroscopic view. 
They traverse the manifold space rapidly and robustly, easily capturing global semantics and structural priors. 
However, due to their large patch size and coarse granularity, their generation trajectory hovers above the manifold. 
They cannot reach the precise image manifold, resulting in structurally correct but blurry images ($x_{coarse}$).
Small patches possess the "fine-grained potential" required to reach the true image manifold. 
However, due to the lack of semantic guidance, they can easily get lost in the vast and complex pixel space. 
Without semantic anchors, fine-grained generation struggles to converge to semantically coherent nodes, often leading to artifact or hallucinative local textures ($x_{fine}$).

To resolve this dilemma and provide explicit semantic anchors for fine-grained generation, we propose HyperDiT, a unified framework that bridges the macroscopic semantic and microscopic pixel manifold via cross-scale interactions.
In HyperDiT, the small-patch sequence is passed through the interaction path, known as Hyper Connector, under the guidance of macroscopic semantics.
Through the Hyper Connectors, the small patchified tokens dynamically query the multi-level semantics across multiple blocks. 
This design ensures that the Fine-grained Flow continuously receives stable semantic guidance from the Semantics Flow throughout the network depth. 
It prevents high-frequency features from deviating from the macroscopic structure, precisely guiding the entire generation trajectory to land on the high-fidelity image manifold ($x_{0}$).

To facilitate the multi-scale Hyper Connectors, we address two critical challenges regarding geometric and feature alignment. 
First, to resolve the inherent spatial mismatch between tokens of different scales, we introduce SA-RoPE. 
This mechanism dynamically scales the rotational basis, ensuring precise geometric alignment across varying resolutions during cross-attention. 
Second, relying solely on large patches for dense cross-scale queries introduces severe semantic ambiguity, as these tokens are heavily entangled with diffusion noise and positional priors. 
To overcome this representation bottleneck, we explore register tokens to extract dense semantics, propagating them through the Semantics Flow and Hyper Connectors. 
Diverging from their conventional use as redundant activation absorbers \cite{darcet2023vision}, we repurpose these tokens as dedicated semantics. 
By explicitly aligning them with a VFM \cite{oquab2023dinov2} during training, these registers capture noise-free dense semantics, thereby providing the fine-grained generation trajectory with precise semantic guidance.

In summary, our main contributions are as follows:
\begin{itemize}
    \item We propose HyperDiT, a pixel-space diffusion framework that addresses the granularity dilemma. We design the small patchified branch as Hyper Connectors, successfully bridging semantic anchors and fine-grained tokens through dense cross-scale interactions.
    \item We introduce SA-RoPE to solve the multi-scale position alignment problem, and incorporate Register Tokens as explicit carriers of dense semantics to provide pure and stable guidance for the fine-grained flow.
    \item Extensive experiments demonstrate that HyperDiT utilizes semantic information to guide fine-grained generation in pixel space, achieving SoTA FID of 1.56 for HyperDiT-H and 1.63 for HyperDiT-XL on ImageNet $256\times256$ and establishing a superior paradigm for high-fidelity image synthesis.
\end{itemize}

\section{Related Work}

\subsection{Latent and Pixel-Space Diffusion Models}
Diffusion models have emerged as the dominant paradigm for high-fidelity image generation \cite{ho2020denoising, song2020denoising}. 
To mitigate the substantial computational overhead of full-image modeling in high-dimensional pixel space using U-Net \cite{ronneberger2015u} or Transformers \cite{peebles2023scalable}, Latent Diffusion Models (LDMs) \cite{rombach2022high} introduced pre-trained VAEs to compress images into a low-dimensional latent space for denoising. 
However, subsequent studies indicate that the inherent information bottleneck of VAEs leads to an irreversible loss of high-frequency details and structural artifacts \cite{chen2025dip, yao2025reconstruction}, ultimately bounding the upper limit of generation quality.

Recently, driven by architectural optimizations and increased computational capacity, the potential of direct pixel-space generation has been re-explored. 
For instance, LlamaGen \cite{sun2024autoregressive} and JiT \cite{jit} demonstrated that highly competitive generation performance can be achieved in pixel space without VAEs, relying solely on standard single-scale Transformers. 
Nevertheless, as image resolution increases, single-branch models face a severe "granularity dilemma." 
To alleviate this, DeCo \cite{ma2025deco} proposed a frequency-decoupled, dual-branch pixel diffusion framework. 
Unlike existing works that focus on frequency-domain decoupling, we revisit pixel-space generation from a manifold learning perspective, exploring a cross-scale feature guidance mechanism.

\subsection{Multi-Scale Architectures in Vision.}
Multi-scale feature representation is a fundamental subject in computer vision. 
During the Convolutional Neural Network (CNN) era, architectures such as FPN \cite{lin2017feature} and HRNet \cite{wang2020deep} established the standard paradigms for multi-resolution feature fusion. 
In the Vision Transformer (ViT) \cite{dosovitskiy2020image} era, works including CrossViT \cite{chen2021crossvit}, MViT \cite{fan2021multiscale}, and Swin Transformer \cite{liu2021swin} further integrated hierarchical and multi-scale designs into attention mechanisms.

Recent advancements in diffusion models have also adopted multi-scale architectures to balance global semantics and local details. 
Models like PixelDiT \cite{yu2025pixeldit} have explored attention mechanisms for small patchified sequences. 
However, it directly compresses pixel tokens before attention, which inevitably leads to a loss of fine-grained information.
DeCo \cite{ma2025deco} employs adaptive layer norm (AdaLN) to incorporate semantics for the high-frequency branch. 
To achieve deep alignment between semantic anchors and pixel-level details, we construct an asymmetric multi-scale diffusion architecture equipped with dense cross-attention mechanisms.

\subsection{Position Encoding and Register Tokens}

\noindent\textbf{Position Encoding.} 
RoPE \cite{su2024roformer} has become standard in foundational models due to its superior relative position awareness and extrapolation capabilities, and it has been successfully extended to 2D vision tasks \cite{heo2024rotary, dehghani2023scaling}. 
However, existing 2D RoPE schemes are primarily designed for single resolutions and cannot directly handle the inherent spatial geometric mismatch between tokens of different scales in a dual-branch network. 
Therefore, designing a scale-aware position encoding that adapts to resolution variations is a critical prerequisite for multi-scale attention interactions.

\noindent\textbf{Register Tokens.} 
While registers were initially introduced in ViTs \cite{darcet2023vision} to absorb redundant activations and mitigate "attention sinks" \cite{xiao2023efficient}, their potential as active information aggregators remains underexplored in generative models. 
Concurrent representation alignment techniques (\textit{e.g.}, REPA \cite{yu2024representation}) have demonstrated that supervising diffusion models with VFMs like DINOv2 \cite{oquab2023dinov2} significantly enhances the semantic richness of the generated features. 
Building upon these insights, we repurpose registers as explicit carriers of dense semantics by aligning these special tokens with pre-trained VFM representations.

\section{Preliminaries}

\noindent \textbf{Flow Matching.} Flow Matching (FM) \cite{lipman2022flow, liu2022flow, holderrieth2025introduction} offers an explicit, continuous-time framework for generative modeling by directly defining a vector field between a simple Gaussian prior and the complex data distribution. 
Unlike traditional diffusion models that rely on discretized Markov chains, FM constructs a direct trajectory $z_t$ that interpolates between noise and data. 
Let $x_0$ denote the clean data sample and ${\epsilon}\sim\mathcal{N}(0,I)$ represent the standard Gaussian noise. 
The latent representation at continuous time step $t\in[0,1]$ is defined via linear interpolation:
\begin{equation}
    {z}_t=tx_0+(1-t)\epsilon
\end{equation}
This formulation guarantees a smooth, continuous path where the state $z_t$ transforms smoothly from pure noise at $t=0$ to the exact data sample at $t=1$. T
he generative dynamics are governed by a target velocity field, computed as the time derivative of this trajectory:
\begin{equation}
    \frac{d{z}_t}{dt}=x_0-\epsilon
\end{equation}
The neural network is trained to approximate this continuous flow. 
The optimization objective minimizes the mean squared error between the model's predicted velocity $v_\theta(z_t,t)$ and the exact derivative:
\begin{equation}
    \mathcal{L}_{FM}(\theta)=\mathbb{E}_{t,{z}_0,{\epsilon}}\left[\|v_\theta({z}_t,t)-({z}_0-\epsilon)\|_2^2\right]
\end{equation}

\noindent \textbf{Classifier-Free Guidance.} Instead of relying on an external pre-trained classifier \cite{dhariwal2021diffusion}, Classifier-Free Guidance (CFG) \cite{ho2022classifier} extrapolates the generative trajectory by jointly learning an unconditional velocity prediction $v_{\theta}({z}_{t}, t, \emptyset)$ and a conditional velocity prediction $v_{\theta}({z}_{t}, t, c)$. 
During inference, the guided velocity field $\tilde{v}_{\theta}({z}_{t}, t, c)$ used for the ODE solver is formulated as:
\begin{equation}
    \tilde{v}_{\theta}({z}_{t}, t, c) = v_{\theta}({z}_{t}, t, \emptyset) + w \cdot \left( v_{\theta}({z}_{t}, t, c) - v_{\theta}(z_{t}, t, \emptyset) \right)
\end{equation}
where $c$ denotes the condition, $\emptyset$ represents a null condition, and $w \ge 1$ is the guidance scale.
By scaling the velocity difference, CFG effectively pushes the target vector field towards regions of higher conditional likelihood.


\section{Methodology}

\begin{figure}[t]
  \centering
  \begin{subfigure}{0.28\textwidth}
    \includegraphics[width=\textwidth]{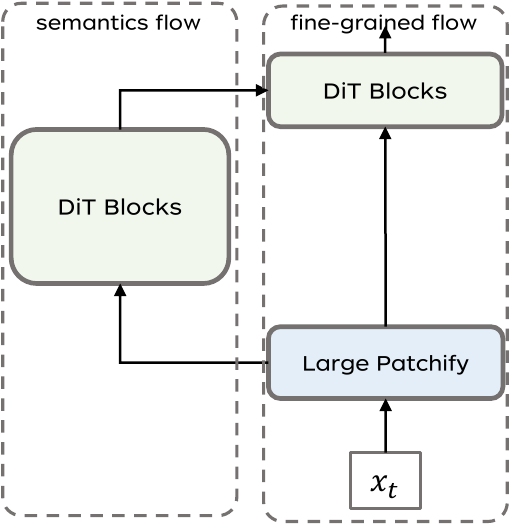}
    \caption{DDT}
    \label{fig:ddt}
  \end{subfigure}
  \hfill
  \begin{subfigure}{0.28\textwidth}
    \includegraphics[width=\textwidth]{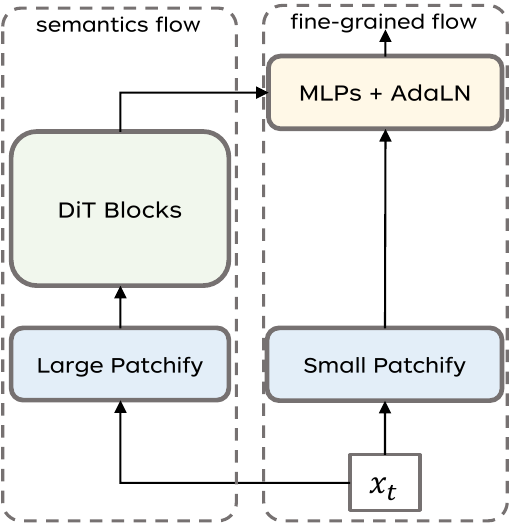}
    \caption{DeCo}
    \label{fig:deco}
  \end{subfigure}
  \hfill
  \begin{subfigure}{0.28\textwidth}
    \includegraphics[width=\textwidth]{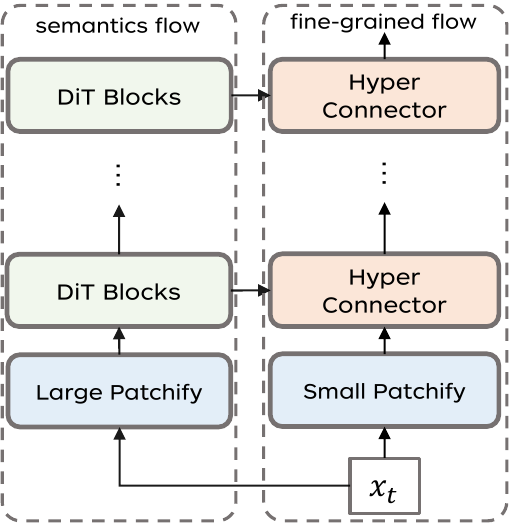}
    \caption{\textbf{HyperDiT (Ours)}}
    \label{fig:hyper}
  \end{subfigure}
  \caption{Architecture comparison. (a) DDT \cite{wang2025ddt}: both semantics and fine-grained flow are processed in large patch size. (b) DeCo \cite{ma2025deco}: the fine-grained flow process semantics through AdaLN layer. (c) HyperDiT: multi-level semantic anchors are transmitted via Hyper Connectors.}
  \label{fig:compare}
\end{figure}

As illustrated in \cref{fig:compare}, many existing multi-scale paradigms process tokens at uniform patch size \cite{wang2025ddt} or employ pixel-wise module (\textit{e.g.}, AdaLN layer) for inter-stream communication \cite{ma2025deco, yu2025pixeldit}, causing suppression on high-frequency semantics.
In our method, HyperDiT propagates semantics by multi-level cross-scale connections (in \cref{sec:hcdit}), each of which carries various dense semantics (in \cref{sec:reg}) and thus is known as Hyper Connector (HC).
This architecture ensures that fine-grained image generation is progressively anchored to multiple semantics.

\subsection{Cross-Connected DiT}
\label{sec:hcdit}

\begin{figure*}
    \centering
    \includegraphics[width=\textwidth]{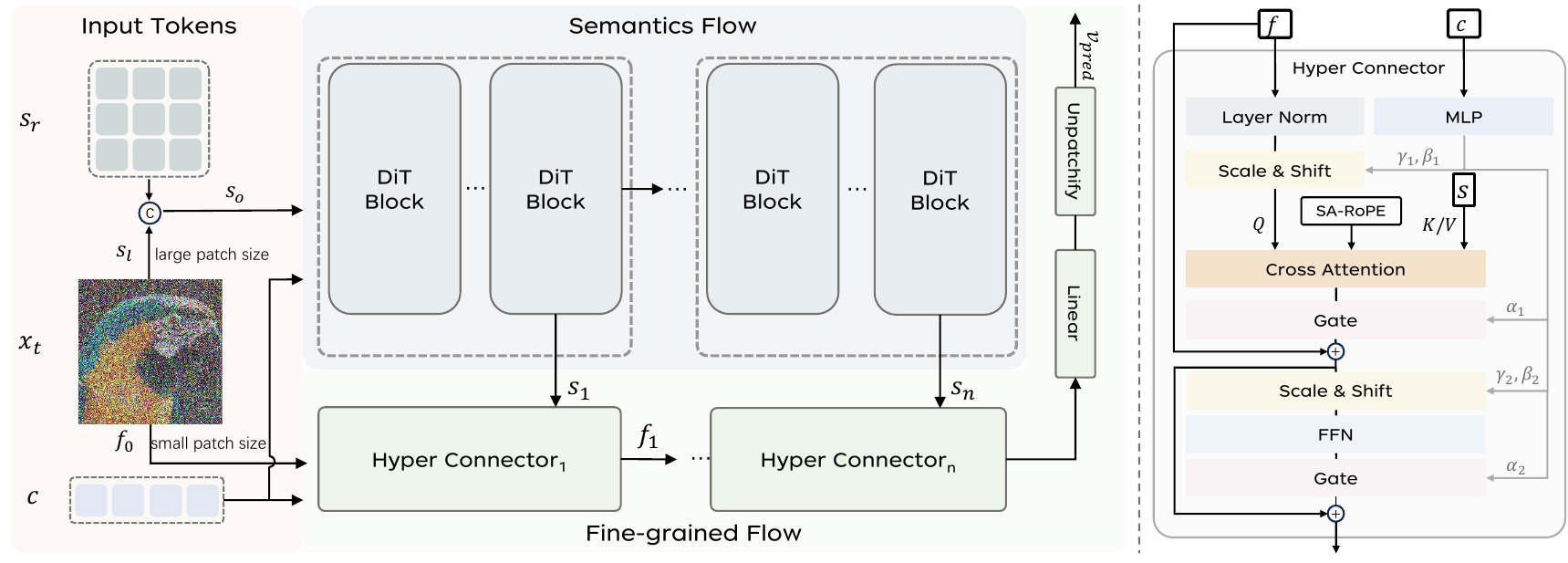}
    \caption{The architecture of HyperDiT. The framework processes global semantics and fine-grained pixels simultaneously. Dense cross-scale interactions are established via Hyper-Connectors, injecting multi-level semantic features into the fine-grained stream. 
    Register tokens $x_{reg}$ are incorporated to capture and propagate the pre-trained semantic feature. In Hyper-Connector, SA-RoPE is used for cross-scale position embedding.}
    \label{fig:framework}
\end{figure*}

The overall architecture, detailed in \cref{fig:framework}, processes the input noisy image $x_t$ in two distinct streams: the Semantics Flow with large patch size and the Fine-grained Flow with small patch size. 
The Semantics Flow provides hierarchical semantic guidance via stacked DiT blocks.
Let $s_{0}$ denote the embedded input tokens of the Semantics Flow.
We select $n$ output responses of DiT blocks to represent the semantic anchors, each of which is denoted as $s_i$ and $i \in \{1, \dots, n\}$.
The learning details of $s_i$ are introduced in \cref{sec:reg}.



The Fine-grained Flow consists of $n$ sequential blocks, and $i$-th block attends the fine-grained queries to the corresponding semantic anchor $s_i$ via Cross-Attention (CA) mechanism.
The structure of each block in Fine-grained Flow, named Hyper-Connector, is illustrated in the right part of \cref{fig:framework}.
Specifically, the $i$-th Hyper-Connector inputs the fine-grained tokens $f_{i-1}$ to generate the query $Q = W_Q \cdot \text{AdaLN}(f_{i-1})$, and the semantic anchor $s_i$ is transferred into key $K = W_K s_i$ and value $V = W_V s_i$.
To resolve the spatial mismatch between tokens of different scales during Cross-Attention, we design a position embedding method, called SA-RoPE, to process $Q$ and $K$ respectively.
The fine-grained features are then updated as:
\begin{equation}
    f_{i} = f_{i-1} + \alpha_1 \cdot \text{CA}(Q, K, V)
\end{equation}
where $\alpha_1$ is the gating parameter from the condition embedding $c$. 
In this way, the generation of fine-grained stream would be progressively guided by the multi-level of semantic anchors.

To visually depict stream-specific features, we adopt a PCA-based visualization method (detailed in Appendix) to visualize the features of Semantics Flow and Fine-grained Flow in \cref{fig:flow}.
The visualization result of Semantics Flow shows smooth background and clear objectiveness regions, indicating the learning of semantic information.
Thanks to semantic anchors, the Fine-grained Flow specializes in capturing structural boundary and local texture simultaneously. 

Our overall training objective is formulated as:
\begin{equation}
    \mathcal{L} = \mathcal{L}_{FM} + \lambda_1 \mathcal{L}_{FreqFM} + \lambda_2 \mathcal{L}_{REPA}
\end{equation}
where $\mathcal{L}_{FreqFM}$ is the frequency FM loss \cite{ma2025deco} and $\mathcal{L}_{REPA}$ is a feature alignment loss \cite{yu2024representation}.
The loss weights are empirically set to $\lambda_1 = 1.0$ and $\lambda_2 = 0.5$.


\subsection{Scale-Aware RoPE}
\label{sec:rope}

\begin{wrapfigure}[23]{r}{0.45\textwidth}
    \centering
    \includegraphics[width=0.45\textwidth]{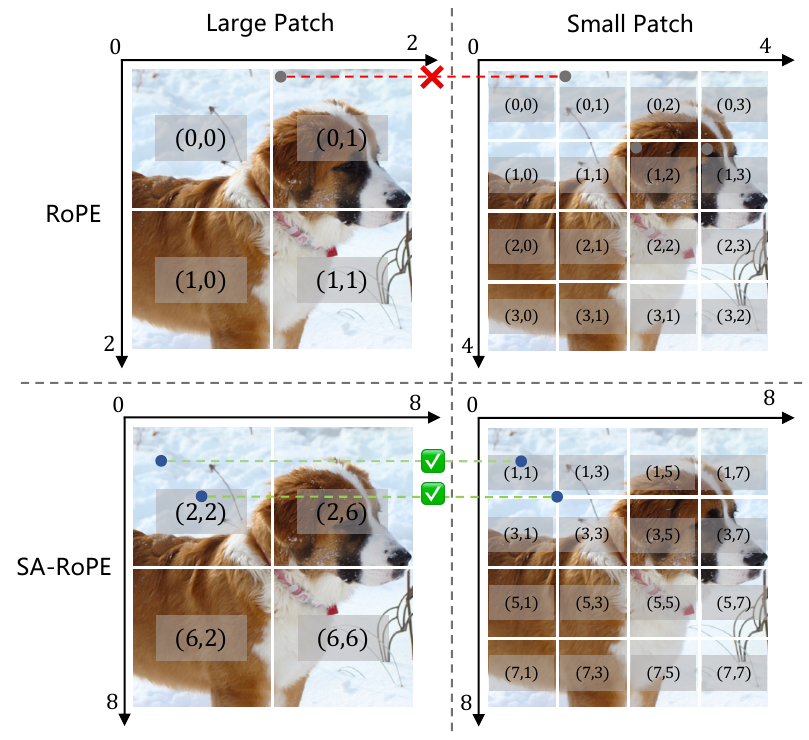}
    \caption{Standard RoPE uses independent grid indices for different patch sizes, which ignores their physical positions. The proposed SA-RoPE ($p_{base}=8$) unifies large and small patches into a shared coordinate and uses center point as position index.}
    \label{fig:rope}
\end{wrapfigure}

In Hyper-Connector, the semantic tokens and fine-grained tokens are generated at different scales.
This cross-scale Cross-Attention requires precise spatial alignment among tokens, where the direct usage of standard RoPE \cite{su2024roformer} leads to critical mismatch. 
Generally, standard 2D RoPE assigns position embeddings based on discrete grid indices of image tokens in raster scan order without considering its visual content.
Therefore, a large patch and small patch of one image may share the same index, even if the content of these two patches are totally different.
As shown in the first row of \cref{fig:rope}, position index $(0,1)$ of the same image represent different regions with different patch sizes.
As the position index impacts the embedded feature of image tokens, we claim that the spatial mismatch may introduce information conflict in cross-scale attentions.

To resolve this mismatch, we introduce SA-RoPE as shown in the second row of \cref{fig:rope}.
SA-RoPE unifies the position embedding of tokens of large patch size $p_l$ and small patch size $p_s$ in a shared coordinate space.
Specifically, we define a base patch size $p_{base}$ to split the image of size $(H,W)$ into a grid of 2D position indices.
After that, the multi-scale tokens from two streams are interleaved within this new grid.
With these constraints, let $L=max(H,W)$ and we define $p_{base}=2^n$ where $n=\lfloor log_2(L/(L/p_s + L/p_l)) \rfloor$.

To calculate the patch index in new grid, we define our task as index transition from original grid of patch size $p \times p$ into new grid, where $p \in \{p_s, p_l\}$.
Given a patch of size $p \times p$ located at the original grid index $(i, j)$, the center point of this patch in the pixel space is $(i \cdot p + p/2, j \cdot p + p/2)$.
Please note that the original grid index $(i, j)$ is calculated according to the top-left pixel position within each patch.
We use the center pixel position to calculate the patch index in our new grid, so that the patches of Semantics Flow and Fine-grained Flow can be interleaved without causing any position conflict.
In this way, our proposed SA-RoPE maps the $(i,j)$-th patch of original grid into an unified coordinate as: 

\begin{equation}
    i' = \frac{i \cdot p + p/2}{p_{base}}, \quad j' = \frac{j \cdot p + p/2}{p_{base}}
\end{equation}
During the cross-attention computation, the dot product between the $Q$ and $K$ inherently attenuates based on their relative distance $\Delta = (i'_{query} - i'_{key}, j'_{query} - j'_{key})$.
This ensures that the fine-grained tokens can accurately absorb context from their geometrically corresponding regions in the semantics anchors.

\subsection{Dense Semantic Learning}
\label{sec:reg}
\begin{figure}[t]
    \centering
    
    \begin{minipage}[b]{0.50\textwidth}
        \centering
        \begin{subfigure}[b]{0.48\linewidth}
            \centering
            \includegraphics[width=\textwidth]{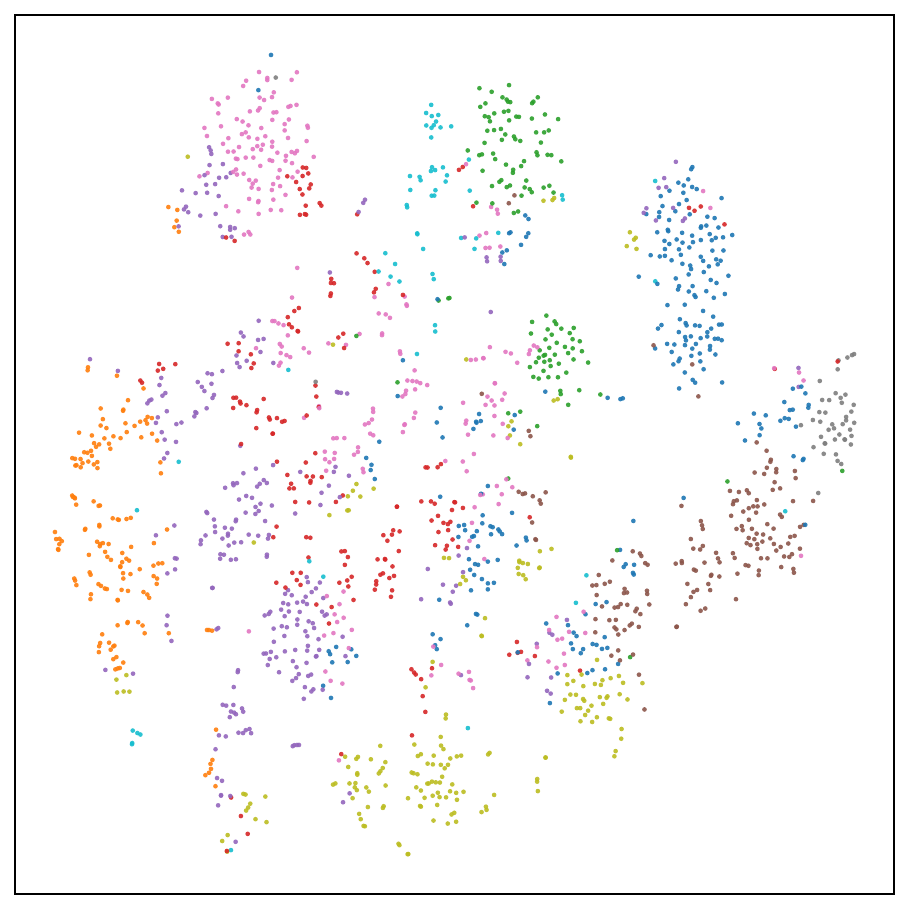}
            \caption{$s_{l}$}
            \label{fig:km_img}
        \end{subfigure}
        \hfill
        \begin{subfigure}[b]{0.48\linewidth}
            \centering
            \includegraphics[width=\textwidth]{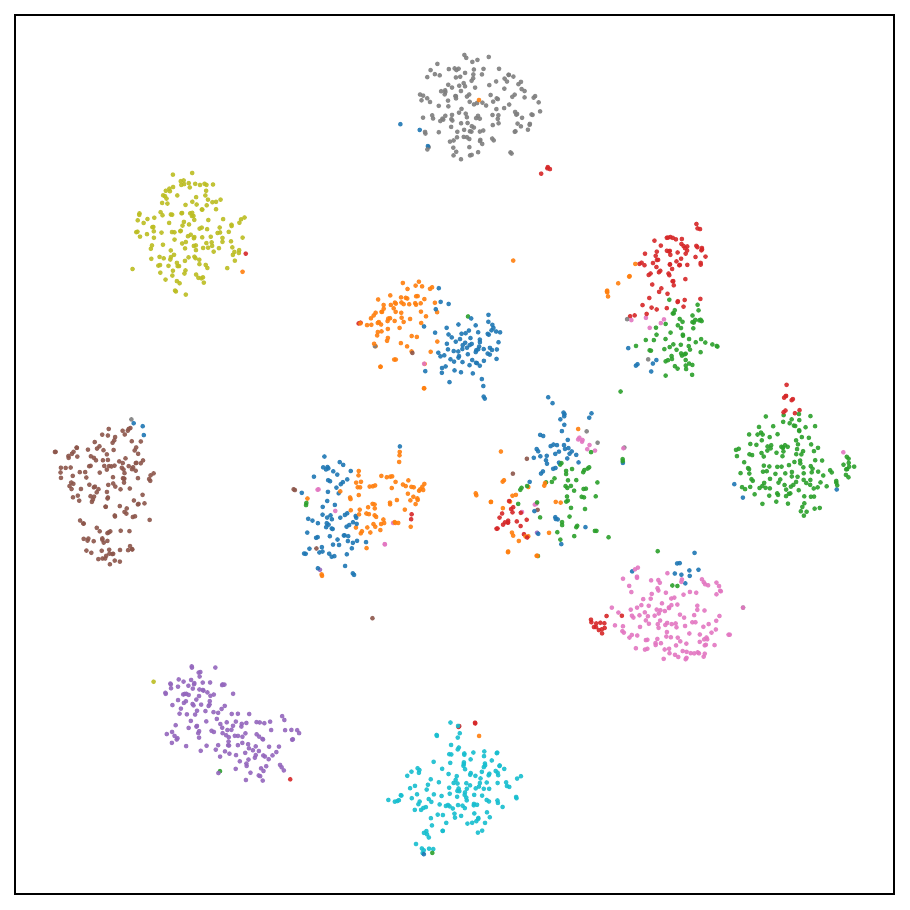}
            \caption{$s_{r}$}
            \label{fig:km_reg}
        \end{subfigure}
        \caption{t-SNE visualization of token embeddings after k-Means (k=10) clustering. (a) Large patchified tokens $s_l$ exhibit entangled distributions. (b) Representation of registers $s_{r}$ forms highly separable clusters.}
        \label{fig:tsne}
    \end{minipage}
    \hfill
    \begin{minipage}[b]{0.46\textwidth}
        \centering
        \includegraphics[width=\linewidth]{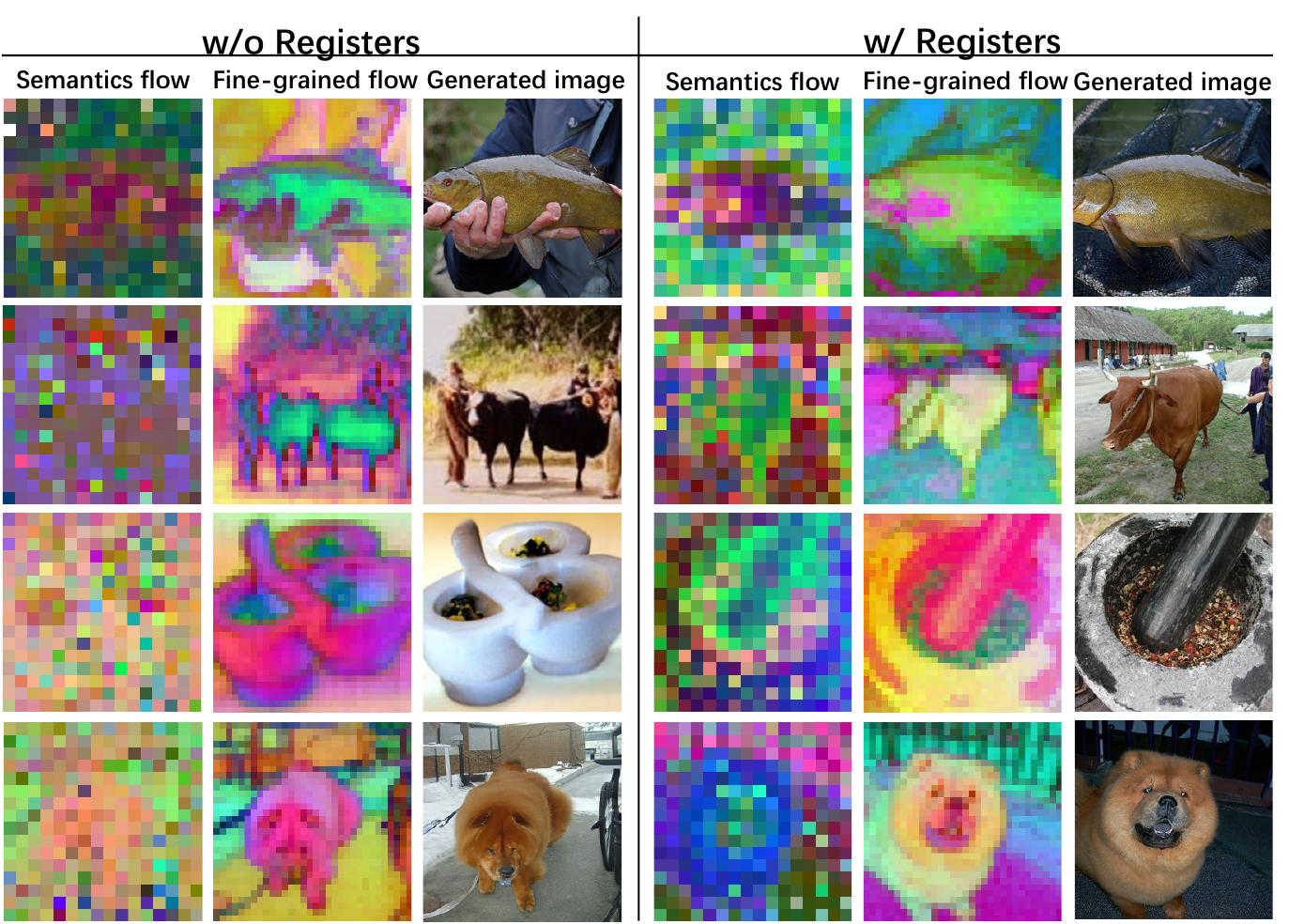}
        \caption{
        The comparison between w/o and w/ registers settings demonstrates that registers enhance the object semantics, thereby mitigating generation hallucination and artifact.
        }
        \label{fig:flow}
    \end{minipage}
    
\end{figure}



In dual-stream diffusion model, the Semantics Flow generates low-frequency semantics while the Fine-grained Flow completes the image details according to the semantics.
Many existing methods \cite{wang2025ddt, ma2025deco,yu2025pixeldit} integrate pre-trained semantic features to guide the learning of large patchified tokens.
Therefore, the large patchified tokens of Semantics Flow stride for a compromise between diffusion noise and pre-trained features, leading to inferior semantic representation.
To tackle this challenge, we input a set of learnable registers $s_r$ along with $s_l$ into Semantics Flow for semantic learning.
To eliminate the impact of position prior, the register tokens do not use any position embeddings.
Therefore, the final input of Semantics Flow is $s_0 = concate(s_r, s_l)$.
In the training process, the semantic features of DINOv2 \cite{oquab2023dinov2} is leveraged to supervise the learning of the representation of register tokens.
Note that we do not constrain the representation distribution of noised patches in Semantics Flow, resulting to the decoupling of semantics for representation and generation.



By aligning the representation of registers with DINOv2 features, the diffusion model learns to encapsulate dense semantic concepts (\textit{e.g.}, distinct object parts or categorical patterns) more than coarse representation.
To validate the representation specialization of tokens, we visualize the token embeddings from Semantics Flow using t-SNE. 
As shown in \cref{fig:tsne}, the representation of large patchified tokens $s_l$ exhibits entangled distributions. 
In contrast, the representation of registers $s_{r}$ form highly separable clusters. 
This indicates that the registers successfully learn and preserve distinct representation semantics from the generation framework. 
We claim that this decoupled design provides the critical advantage that the pre-trained representation semantics can be fully integrated in diffusion model.
The result in \cref{fig:flow} demonstrates that the incorporation of the semantic-aligned representation of registers reduces visual hallucinations and artifacts, preventing texture collapse and maintaining superior structural coherence in the generated images.

\section{Experiments}
\begin{table}[t]
\centering
\caption{Quantitative comparison on ImageNet $256\times 256$ with CFG. All metrics are evaluated using 50K samples. "Params" reports the inference parameter count (generator + VAE/RAE decoder). NFE: Number of Function Evaluations.}
\label{tab:compare}
\resizebox{\textwidth}{!}{%
\begin{tabular}{l c c c c c c c c}
\toprule
\textbf{ImageNet 256$\times$256} & \textbf{Params} & \textbf{Epochs} & \textbf{NFE} & \textbf{FID}$\downarrow$ & \textbf{IS}$\uparrow$ & \textbf{sFID$\downarrow$} & \textbf{Pre.$\uparrow$} & \textbf{Rec.$\uparrow$} \\
\midrule
\rowcolor{gray!20} \multicolumn{9}{l}{\textit{Latent-space Diffusion}} \\
    DiT-XL/2 \cite{peebles2023scalable} & 675+49M & 1400 & 250$\times$2 & 2.27 & 278.2 & 4.60 & \textbf{0.83} & 0.57 \\
    SiT-XL/2 \cite{ma2024sit} & 675+49M & 1400 & 250$\times$2 & 2.06 & 277.5 & 4.50 & 0.80 & 0.64 \\
    REPA-XL/2 \cite{yu2024representation} & 675+49M & 800 & 250$\times$2 & 1.42 & 305.7 &4.70 & 0.80 & 0.64 \\
    LightningDiT-XL/2 \cite{yao2025reconstruction} & 675+49M & 800 &250$\times$2 & 1.35 & 295.3 & \textbf{4.15} & 0.79 & 0.65 \\
    DDT-XL/2 \cite{wang2025ddt}  & 675+49M & 400 & - & 1.26 & \textbf{310.6} & - & 0.78 & \textbf{0.67}\\
    RAE-XL/2 \cite{zheng2025diffusion} & 839+415M & 800 & - & \textbf{1.13} & 262.6 & - & 0.78 & \textbf{0.67} \\
\midrule
\rowcolor{gray!20} \multicolumn{9}{l}{\textit{Pixel-space (non-diffusion)}} \\
    JetFormer \cite{tschannen2024jetformer} & 2.8B & - & - & 6.64 & - & - & 0.69 & 0.56 \\
    FractalMAR-H \cite{li2025fractal} & 848M & 800 & - & 6.15 & 348.9 & - & 0.81 & 0.46 \\
\midrule
\rowcolor{gray!20} \multicolumn{9}{l}{\textit{Pixel-space Diffusion}} \\
    ADM-G \cite{dhariwal2021diffusion} & 554M & 400 & 250 & 4.59 & 186.7 & 5.25 & 0.82 & 0.52 \\
    RIN \cite{jabri2022scalable} & 410M & 480 & - & 3.42 & 182.0 & - & - & - \\
    SiD \cite{hoogeboom2023simple} & 2B & - & - & 4.08 & 256.3 & - & - & - \\
    PixelFlow-XL/4 \cite{chen2025pixelflow} & 677M & 320 & 120$\times$2   & 1.98 & 282.1 & 5.83 & 0.81 & 0.60 \\
    PixNerd-XL/16 \cite{wang2025pixnerd} & 700M & 320 & 100$\times$2 & 1.95 & 300 & 4.54 & 0.80 & 0.60 \\
    JiT-G/16 \cite{jit} & 2B & 600 & 100$\times$2 & 1.82 & 292.6 & - & - & - \\
    PixelGen-XL/16 \cite{ma2026pixelgen} & 676M & 160 & 100$\times$2 & 1.83 & 293.6 & - & 0.79 & 0.63 \\
    DiP-XL/16 \cite{chen2025dip} & 631M & 600 & 100$\times$2 & 1.79 & 281.9 & 4.59 & 0.80 & 0.63 \\
    DeCo-XL/16 \cite{ma2025deco} & 682M & 600 & 100$\times$2 & 1.69 & 304.0 & 4.59 & 0.79 & 0.63 \\
\midrule
    \rowcolor{green!10} \textbf{HyperDiT-XL} & 676M & 600 & 100$\times$2 & 1.63 & 304.2 & 4.76 & 0.80 & 0.62 \\
    \rowcolor{green!10} \textbf{HyperDiT-H} & 952M & 600 & 100$\times$2 & 1.56 & 306.5 & 4.73 & 0.80 & 0.64 \\
\bottomrule
\vspace{-15pt}
\end{tabular}
}
\end{table}

\subsection{Implementation Details}
We denote our models as HyperDiT-$X$, where $X$ indicates the model size (\textit{e.g.}, XL).
We conduct experiments on $256 \times 256$ ImageNet dataset. 
The large patch size, small patch size, and base patch size are set to 16, 8 and 4 respectively.
The model is optimized using the Adam optimizer with a learning rate of $5 \times 10^{-5}$ and a total batch size of 1024.
All models are trained using 8 B200 GPUs. 
During inference, we use the Heun sampler \cite{heun1900neue} with 50 sampling steps by default. 
We employ FID \cite{heusel2017gans}, sFID \cite{nash2021generating}, IS \cite{salimans2016improved}, Precision and Recall \cite{kynkaanniemi2019improved} as evaluation metrics.

\subsection{Comparison Results}

\begin{figure}[t]
    \centering
    \begin{subfigure}{0.49\textwidth}
        \centering
        \includegraphics[width=\textwidth]{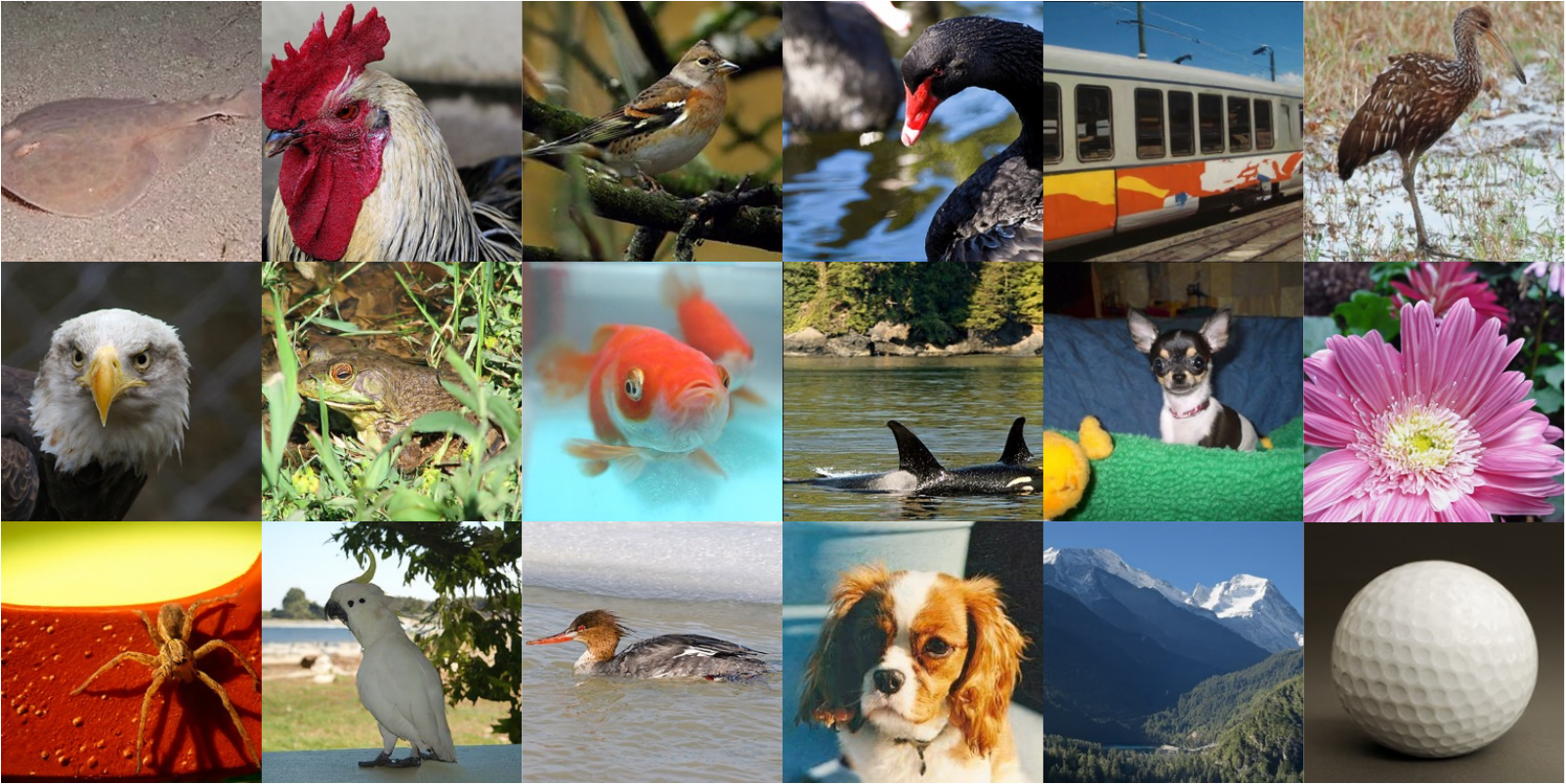}
        \caption{HyperDiT-XL}
        \label{fig:demo_xl}
    \end{subfigure}
    \hfill
    \begin{subfigure}{0.49\textwidth}
        \centering
        \includegraphics[width=\textwidth]{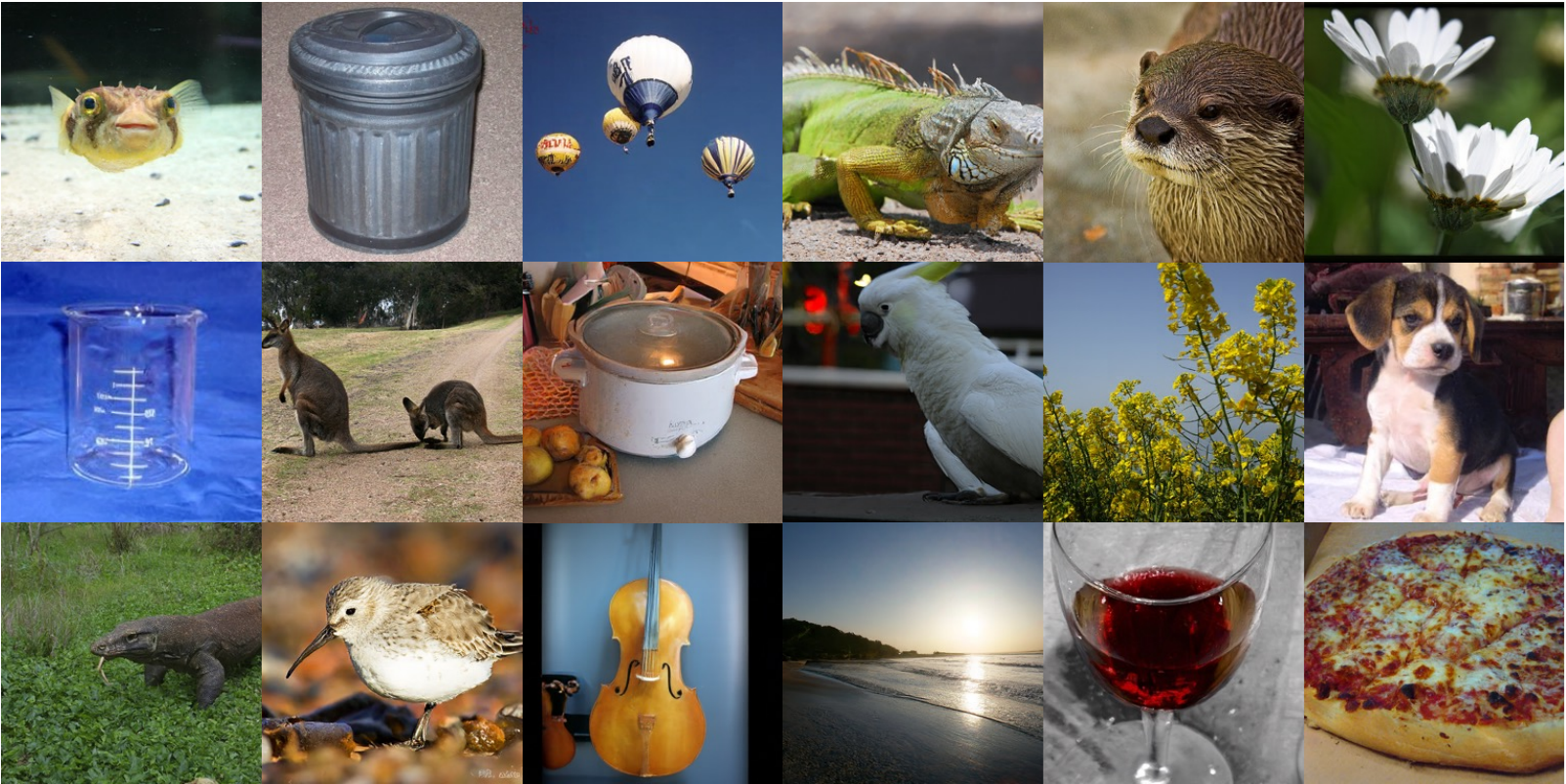}
        \caption{HyperDiT-H}
        \label{fig:demo_h}
    \end{subfigure}

    \caption{Visualization of the generated images by HyperDiT-XL and HyperDiT-H at $256\times256$ resolution. More qualitative results can be found in Appendix \ref{appendix:more_visulized_results}.}
    \label{fig:demo_xl_h}
    \vspace{-10pt}
\end{figure}

As summarized in \cref{tab:compare}, HyperDiT achieves SoTA performance in pixel-space generation. 
By directly modeling the pixel distribution, HyperDiT-H achieves an FID of 1.56, significantly outperforming recent strong baselines such as JiT-G/16 (FID 1.82) \cite{jit} and DiP-XL/16 (FID 1.79) \cite{chen2025dip}. 
Notably, our HyperDiT-XL (FID 1.65) surpasses DeCo-XL/16 (FID 1.69) \cite{ma2025deco}. 
In addition, HyperDiT bridges the performance gap with latent-space models without relying on VAE or RAE \cite{zheng2025diffusion}. 
Both HyperDiT-XL and HyperDiT-H outperforms foundational latent models, including DiT-XL/2 (FID 2.27) \cite{peebles2023scalable} and SiT-XL/2 (FID 2.06) \cite{ma2024sit}.
Beyond FID, HyperDiT-H achieves a competitive IS of 306.5 and a Precision of 0.80. 
The visualization results are demonstrated in \cref{fig:demo_xl} and \cref{fig:demo_h}.

\subsection{Ablation Study}

\subsubsection{Setup.}
We conduct ablation studies on the ImageNet dataset at a resolution of $256 \times 256$. 
Unless otherwise specified, all ablation models adopt the HyperDiT-XL architecture and are trained from scratch for 50 epochs. 
During evaluation, images are generated using the Heun \cite{heun1900neue} sampler with 50 inference steps. 
CFG is set to 3.2. 
The CFG intervals are set to $CFG_{min} = 0.1$ and $CFG_{max} = 1.0$.


\begin{table}[h]
  \centering
  \caption{Ablation studies on model design choices. (a) Comparison of different patch sizes in HyperDiT. (b) Effect of the number of HCs. (c) Effect of the number of registers $l$.}
  \label{tab:main_ablations}
  \scriptsize
  \setlength{\tabcolsep}{3pt}

  \begin{subtable}[b]{0.31\textwidth}
    \centering
    \resizebox{\linewidth}{!}{%
    \begin{tabular}{l|c|c|c|c}
      \toprule
      Exp & $p_l$-$p_s$  & FID $\downarrow$ & IS $\uparrow$ \\
      \midrule
      1 & 32-16   & 112.51 & 8.67 \\
      2 & 32-8   & 80.31  & 21.94 \\
      3 & 16-16   & 92.34  & 16.58 \\
      \rowcolor{gray!15}4 & 16-8  & 66.28 & 28.61 \\
      \bottomrule
    \end{tabular}%
    }
    \caption{Effect of patch size.}
    \label{tab:patch}
  \end{subtable}
  \hfill
  \begin{subtable}[b]{0.40\textwidth}
    \centering
    \resizebox{\linewidth}{!}{%
    \begin{tabular}{l|c|c|c|c|c}
      \toprule
      Exp & $m$ & $n$ & Params & FID $\downarrow$ & IS $\uparrow$ \\
      \midrule
      1 & 12 & 2 & 628M & 5.98 & 182.2 \\
      2 & 8  & 3 & 652M & 5.19 & 194.9 \\
      \rowcolor{gray!15}3 & 6 & 4 & 676M & 4.27 & 212.7 \\
      4 & 4  & 6 & 724M & 4.08 & 213.6 \\
      \bottomrule
    \end{tabular}%
    }
    \caption{Number of HCs.}
    \label{tab:num_blocks}
  \end{subtable}
  \hfill
  \begin{subtable}[b]{0.24\textwidth}
    \centering
    \resizebox{\linewidth}{!}{%
    \begin{tabular}{l|c|c|c}
      \toprule
      Exp & $l$ & FID $\downarrow$ & IS $\uparrow$ \\
      \midrule
      1 & 0   & 6.22 & 175.1 \\
      2 & 4   & 5.27 & 185.5 \\
      3 & 16  & 4.98 & 193.0 \\
      4 & 64  & 5.35 & 189.9 \\
      \rowcolor{gray!15}5 & 256 & 4.27 & 212.7 \\
      \bottomrule
    \end{tabular}%
    }
    \caption{Number of registers.}
    \label{tab:reg}
  \end{subtable}
\vspace{-10pt}
\end{table}

\subsubsection{Effect of Patch Size.}
We investigate the impact of large ($p_l$) and small ($p_s$) patch sizes using the 131M-parameter HyperDiT-B. 
As shown in \cref{tab:patch}, a overly larger patch size severely limits stable semantic guidance; reducing $p_l$ from 32 to 16 (comparing the 32-16 and 16-16 configurations) decreases the FID from 112.51 to 92.34. 
Furthermore, finer granularity in the fine-grained flow is crucial for capturing high-frequency pixel details. 
By fixing $p_l=16$ and reducing $p_s$ from 16 to 8, the FID drops substantially to 66.28, while the IS surges from 16.58 to 28.61. 
However, scaling to extreme pixel-level granularity ($p_s=4$) inevitably leads to Out-Of-Memory (OOM) errors due to the quadratic complexity of attention over massively extended sequences. 
Consequently, the 16-8 configuration achieves the optimal balance between computational feasibility (43 GFLOPs) and generation fidelity, serving as our default setting.
 

\subsubsection{Number of Hyper Connectors.}

We evaluate the impact of the interaction frequency between the semantics flow and the fine-grained flow, governed by the margin $m$ and the number of HCs $n$. 
With the total number of DiT blocks in the semantics flow fixed at 24, $m$ dictates the interval at which semantic anchors are extracted and passed to the fine-grained branch (\textit{i.e.}, $m \times n = 24$). 
As detailed in \cref{tab:num_blocks}, a configuration with sparse interactions ($m=12, n=2$) yields a sub-optimal FID of 5.98. 
Increasing the interaction density to $n=4$ provides a significant performance boost, decreasing the FID to 4.27. 
This demonstrates that more frequent cross-scale queries are important for anchoring high-resolution features, preventing them from getting lost in the pixel space. 
While further increasing the Hyper Connectors to $n=6$ ($m=4$) marginally decreases the FID to 4.08, it inflates the parameter count to 724M. Consequently, we adopt $m=6$ and $n=4$ as our default configuration, achieving an optimal balance between high-fidelity generation and architectural efficiency.

\subsubsection{Number of Registers.}
We investigate the impact of the number of register tokens $l$ on generation quality. 
As shown in \cref{tab:reg}, operating without registers ($l=0$) yields suboptimal performance.
This confirms that relying solely on large patches for cross-scale queries entangles noise and semantics, leading to ambiguous guidance for the fine-grained flow. 
Introducing even a small number of registers ($l=4, 16$) provides immediate gains by serving as dedicated, noise-free semantic anchors. 
Notably, scaling the number of registers to $l=256$ achieves the most significant performance boost, reaching an optimal FID of 4.27 and an IS of 212.7. 
This improvement indicates that a larger capacity of register tokens is essential to adequately encapsulate diverse, dense local semantic concepts (\textit{e.g.}, distinct object parts and categorical patterns). 
Consequently, these abundant registers deliver robust and detailed guidance to the HC, effectively mitigating visual hallucinations.
Based on these observations, we adopt 256 as the default number of registers.

\subsubsection{Effect of $\mathcal{L}_{REPA}$.}
\begin{wraptable}[10]{r}{0.4\textwidth}
  \centering
  \caption{Effect of $\mathcal{L}_{REPA}$ applied to large patches $s_l$ and registers $s_r$.}
  \label{tab:repa}
  \setlength{\tabcolsep}{5pt}
  \small
  \begin{tabular}{l|c|c|c|c}
    \toprule
    Exp & $s_l$ & $s_r$ & FID $\downarrow$ & IS $\uparrow$ \\
    \midrule
    1 & \ding{55} & \ding{55} & 5.64 & 179.5 \\
    \rowcolor{gray!15}2 & \ding{55} & \ding{51} & 4.27 & 212.7 \\
    3 & \ding{51} & \ding{55} & 4.92 & 191.6 \\
    4 & \ding{51} & \ding{51} & 5.23 & 197.3 \\
    \bottomrule
  \end{tabular}
\end{wraptable}
We investigate the optimal target for the REPA objective $\mathcal{L}_{REPA}$ defined in \cite{yu2024representation}. 
For this ablation, we exclusively vary the alignment target within the semantics flow. 
As shown in \cref{tab:repa}, substituting the register tokens $s_{r}$ with the large patches $x_{l}$ degrades the FID to 4.92. 
This performance drop stems from a severe representation conflict: forcing large patches, which must intrinsically model positional layouts and the denoising trajectory, to simultaneously encode dense semantic features disrupts the optimization process. 
Applying the alignment objective to both $s_{l}$ and $s_{r}$ exacerbates this interference, yielding the worst FID of 5.23. 
In contrast, applying the alignment exclusively to $s_{r}$ achieves the optimal FID of 4.27 and IS of 212.7. 
Because register tokens are non-spatial, they can serve as dedicated semantic anchors without interfering with the denoising process, elegantly decoupling dense semantic understanding from large patchified noised tokens.

\subsubsection{Effectiveness of each component.}

\begin{wrapfigure}[12]{r}{0.45\textwidth}
    \centering
    \setlength{\tabcolsep}{6pt}
    \captionof{table}{Effectiveness of each component.}
    \label{tab:compon}
    \begin{tabular}{l|c|c}
      \toprule
      Method & FID $\downarrow$ & IS $\uparrow$ \\
      \midrule
      Baseline (DeCo) \cite{ma2025deco}   & 8.95 & 156.1 \\
      + Dense Connections     & 7.74 & 162.1 \\
      \quad + Hyper Connectors        & 7.04 & 166.0 \\
      \quad\quad + SA-RoPE           & 6.22 & 175.0 \\
      \quad\quad\quad + Registers       & 5.01 & 192.3 \\
      \rowcolor{gray!15}\quad\quad\quad\quad + $\mathcal{L}_{REPA}$   & 4.27 & 212.7 \\
      \bottomrule
    \end{tabular}
\end{wrapfigure}

We conduct a step-by-step ablation to validate each proposed module, as detailed in \cref{tab:compon}.
We adopt DeCo \cite{ma2025deco} as our baseline, which processes the fine-grained flow using an MLP combined with AdaLN and incorporates semantic guidance from the last level, yielding an initial FID of 8.95.
Integrating Dense Connections, transmitting multi-level semantic anchors rather than a single final output, improves the FID to 7.74.
This confirms that multi-level guidance effectively prevents fine-grained features from diverging from the structure throughout the generation process. 
Subsequently, introducing Hyper Connectors to the Fine-grained Flow further reduces the FID to 7.04. 
This demonstrates that utilizing CA mechanism allows fine-grained tokens to dynamically query semantic anchors, providing far more expressive and accurate feature fusion than AdaLN.
Incorporating SA-RoPE in CA yields a substantial improvement, bringing the FID down to 6.22. 
This highlights the necessity of resolving the inherent spatial mismatch; without SA-RoPE, CA fails to establish accurate position alignment between tokens of varying resolutions. 
Finally, adding Registers pushes the FID to 5.01, and explicitly enforcing their dense semantics via $\mathcal{L}_{REPA}$ achieves the optimal FID of 4.27 and an IS of 212.7. 
This validates that non-spatial registers, when properly supervised, successfully decouple dense semantic understanding from spatial denoising, delivering robust, noise-free local guidance to complete the high-fidelity generation.

\section{Conclusion}

In this work, we presented HyperDiT, a multi-scale diffusion framework designed to overcome the "granularity dilemma" in pixel-space image generation.
The Hyper Connectors are proposed to bridge the semantic and pixel manifolds.
Diverging from prior methods that rely on the AdaLN layer to process cross-scale tokens, HyperDiT leverages dense cross-attention mechanisms, enabling fine-grained tokens to query multi-level semantic anchors throughout the network.
We introduce SA-RoPE to guarantee the position alignment of the cross-scale interactions. 
Additionally, the Registers are repurposed to capture noise-free dense semantics from a VFM, effectively eliminating final generation hallucinations.
Extensive evaluations demonstrate that our approach achieves SoTA performance on ImageNet $256\times256$. 
By seamlessly utilizing semantic anchors to guide fine-grained tokens, HyperDiT establishes a robust and superior paradigm for pixel-space diffusion.

\bibliographystyle{unsrtnat} 
\bibliography{main}
\newpage
\appendix

\section{Additional Implementation Details}

\subsection{Hyperparameters}
\cref{tab:hyper_param} details the configurations for the HyperDiT-B, XL, and H variants. 
Model scaling is primarily driven by expanding the base DiT blocks and hidden dimensions. 
Notably, the number of our proposed Hyper Connectors remains constant at 4 across all scales. 
To ensure strictly fair comparisons during ablation and scaling studies, all training hyperparameters, including epochs, batch size, and learning rate, are kept identical across the three models.
During sampling, the CFG scale is slightly reduced for the larger XL and H models (from 3.2 to 2.9).
\begin{table*}[h]
\centering
\setlength{\tabcolsep}{8pt}
\caption{Experimental configurations for HyperDiT models. We detail the architecture, training, and sampling hyperparameters for the B, XL, and H variants. }
\label{tab:hyper_param}
\resizebox{0.8\textwidth}{!}{
\begin{tabular}{l|c|c|c}
\toprule
 & \textbf{HyperDiT-B} & \textbf{HyperDiT-XL} & \textbf{HyperDiT-H} \\
\midrule
    \rowcolor{gray!20} \multicolumn{4}{l}{\textbf{Architecture}} \\
    DiT blocks & 8 & 24 & 28 \\
    Hyper Connectors & 4 & 4 & 4 \\
    hidden dim & 768 & 1152 & 1280 \\
    large patch size & 16 & 16 & 16 \\
    small patch size & 8 & 8 & 8 \\
    num heads & 16 & 16 & 16 \\
    num registers & 256 & 256 & 256 \\
\midrule
    \rowcolor{gray!20} \multicolumn{4}{l}{\textbf{Training}} \\
    epochs & 600 & 600 & 600 \\
    warmup epochs & 5 & 5 & 5 \\
    batch size & 1024 & 1024 & 1024 \\
    learning rate & 5e-5 & 5e-5 & 5e-5 \\
    learning rate schedule & constant & constant & constant \\
    ema decay & 0.9999 & 0.9999 & 0.9999 \\
    time sampler & lognorm \cite{esser2024scaling} & lognorm & lognorm \\
\midrule
    \rowcolor{gray!20} \multicolumn{4}{l}{\textbf{Sampling}} \\
    ODE solver & Heun \cite{heun1900neue} & Heun & Heun \\
    ODE steps & 50 & 50 & 50 \\
    CFG scale & 3.2 & 2.9 & 2.9 \\
    CFG interval \cite{kynkaanniemi2024applying} & [0.1, 1.0] & [0.1, 1.0] & [0.1, 1.0] \\
\bottomrule
\end{tabular}
}
\end{table*}
\subsection{Effect of CFG.}

\begin{figure}[h]
    \centering
    \includegraphics[width=0.4\textwidth, trim=0pt 0pt 0pt 20pt]{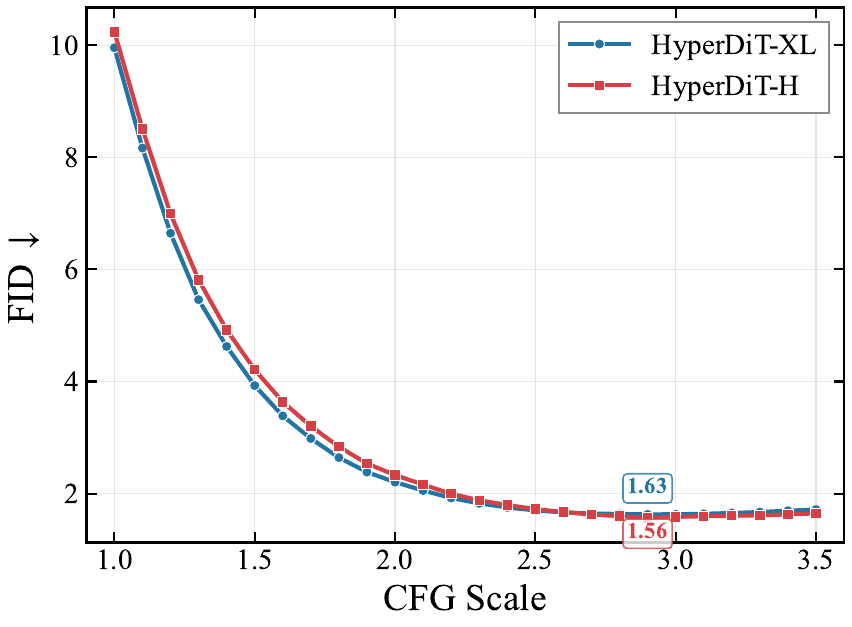}
    \caption{Effect of CFG scale.}
    \label{fig:cfg}
    \vspace{-10pt}
\end{figure} 
We investigate the effect of the CFG scale on generation quality, as illustrated in \cref{fig:cfg}. 
HyperDiT-XL and HyperDiT-H are both trained from scratch for 600 epochs.
When generating images without guidance (CFG = 1.0), the models exhibit relatively high FID scores. 
As the CFG scale increases, the FID steadily decreases. 
The optimal performance is achieved at a CFG scale of 2.9, where HyperDiT-XL and HyperDiT-H reach their minimum FID scores of 1.63 and 1.56, respectively. Further increasing the CFG scale yields no additional benefits.

\subsection{Details of PCA Visualization}

\begin{figure}[t]
    \centering
    \includegraphics[width=\textwidth]{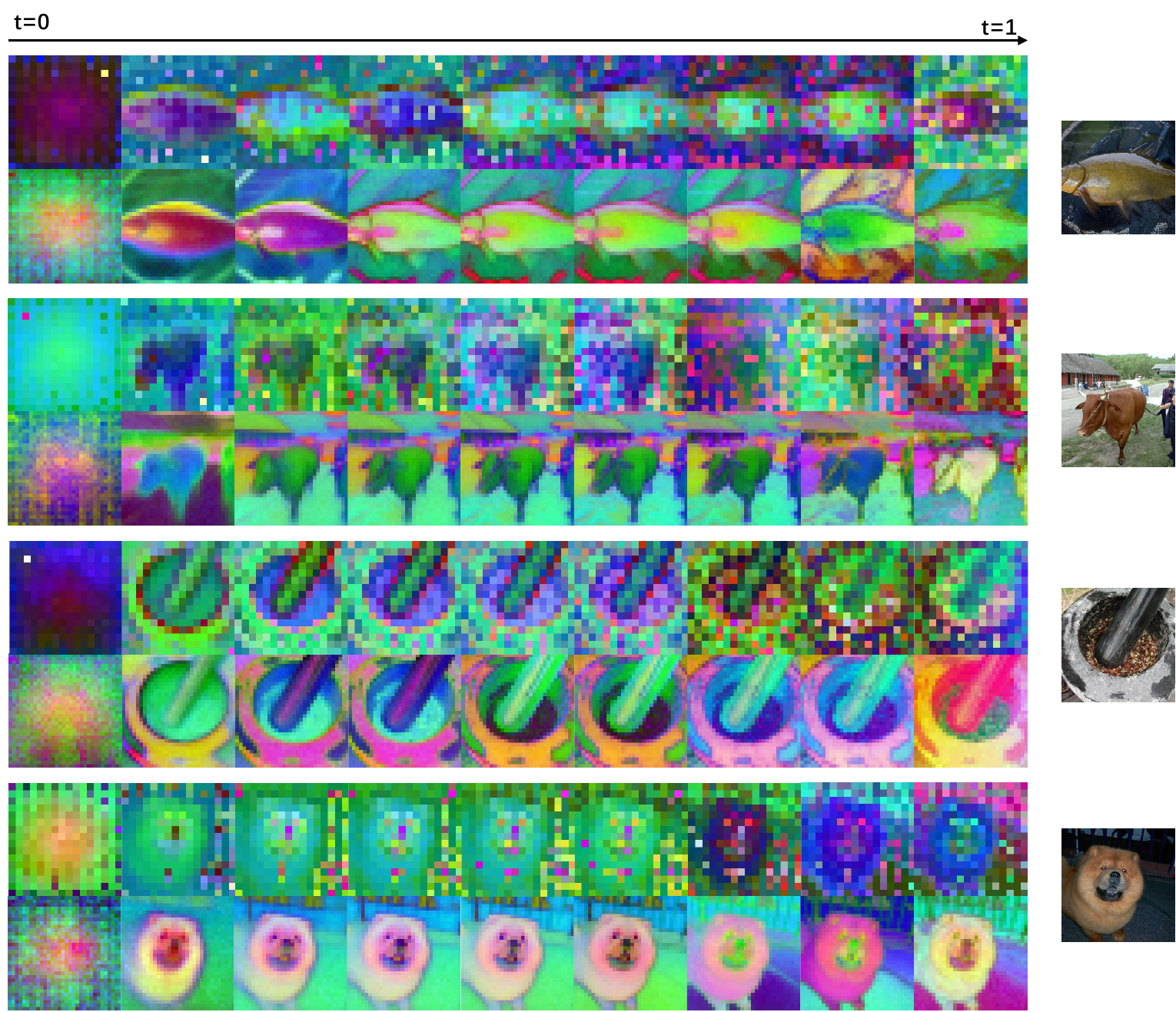}
    \caption{PCA visualization of token embeddings across different timesteps. 
    For each example image shown on the right, the top row visualizes the token embeddings from the Semantics Flow, while the bottom row visualizes those from the Fine-grained Flow.}
    \label{fig:pca_t}
\end{figure}

To intuitively understand the internal representations of the Semantics Flow and Fine-grained Flow, we employ Principal Component Analysis (PCA) to visualize the high-dimensional feature spaces (as shown in \cref{fig:pca_t}). 
The visualization process follows a standard dimensionality reduction pipeline.
During the forward pass of the diffusion process, we extract the token embeddings from the last block of both the semantic and fine-grained flows. 
Let the extracted spatial features be denoted as $X \in \mathbb{R}^{H \times W \times D}$, where $H \times W$ represents the spatial resolution (number of patches) and $D$ is the hidden dimension of the corresponding block.
To visualize the $D$-dimensional features, we flatten the spatial dimensions to construct a 2D feature matrix $X' \in \mathbb{R}^{N \times D}$, where $N = H \times W$. 
We then apply PCA to project $X'$ onto its top three principal components, resulting in a reduced feature matrix $Y \in \mathbb{R}^{N \times 3}$. Next, we apply min-max normalization independently to each of the three principal components to scale their values into the $[0, 1]$ range. Finally, the normalized components are reshaped back to the spatial resolution $H \times W$ and mapped directly to the RGB channels, producing the feature maps shown in our figures.

Notably, as observed in \cref{fig:pca_t}, the visualizations of the semantics flow exhibit a trend of becoming progressively blurrier from $t=0$ to $t=1$. 
This phenomenon intuitively reflects the dynamic division of labor across timesteps of our dual-branch architecture. 
As the generation process advances toward the clean image, the Semantics Flow increasingly specializes in modeling low-frequency (\textit{e.g.}, smooth color transitions) . 
Consequently, the burden of synthesizing sharp edges and complex textures is entirely delegated to the Fine-grained Flow, leading to the smoothed and homogeneous appearance of the semantics features at later timesteps.

\begin{figure}[t]
    \centering
    \includegraphics[width=\textwidth]{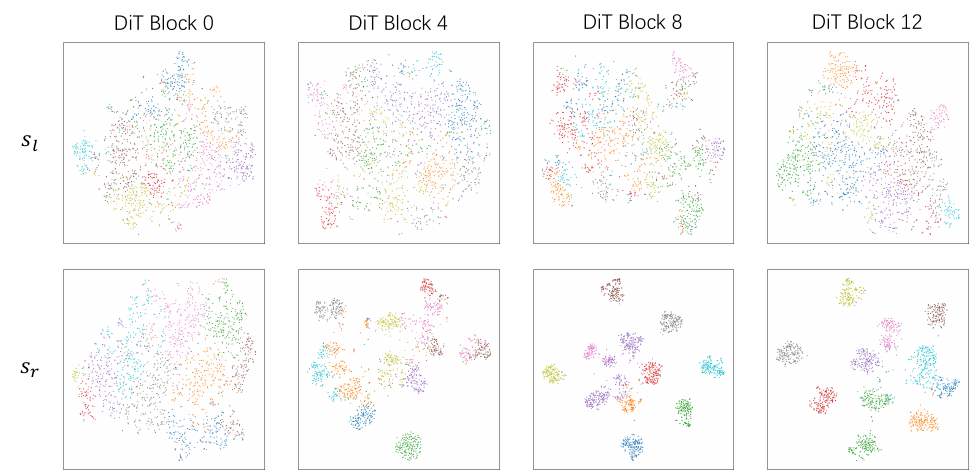}
    \caption{t-SNE visualization of the large patchified tokens $s_l$ and the registers $s_r$ across different Semantics Flow depths (specifically at DiT Blocks 0, 4, 8, and 12).}
    \label{fig:tsne}
\end{figure}

\subsection{Details of t-SNE Visualization}

To provide a deeper understanding of the feature representations learned by our model, we detail the t-SNE visualization process and present additional layer-wise results in \cref{fig:tsne}.
Specifically, we randomly sample $10$ distinct categories in ImageNet, with $160$ images selected per category.
We extract the large patchified tokens $s_l$ and the register tokens $s_r$ from the Semantics Flow. 
To process these high-dimensional features, we first apply K-Means clustering independently to $s_l$ and $s_r$. 
Subsequently, we utilize t-SNE to project these clustered features into a 2D space for visualization.

As illustrated in \cref{fig:tsne}, $s_l$ remain heavily entangled throughout the network depth, as they are inherently coupled with diffusion noise and coarse spatial priors. 
In contrast, while $s_r$ are initially mixed at Block 0, they progressively form highly separable and distinct semantic clusters as the layers deepen. 
This dynamic layer-wise progression provides strong empirical evidence for our claim in the main text: the registers effectively decouple from noise and successfully aggregate dense semantics as the features propagate through the Semantics Flow.

\begin{figure}[t]
    \centering
    
    \begin{minipage}[b]{0.46\textwidth}
        \centering
        \includegraphics[width=\linewidth]{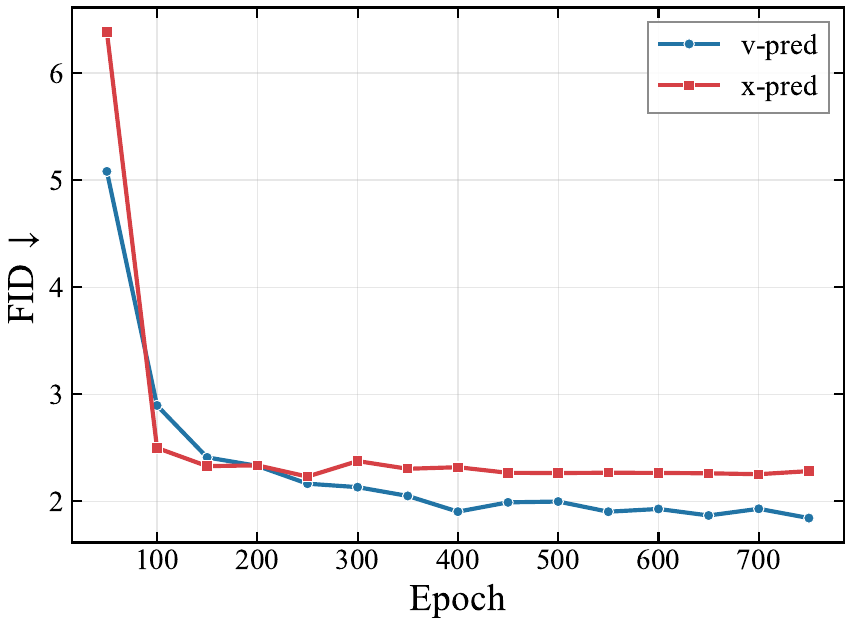}
        \caption{
        FID of $x$-pred and $v$-pred.
        }
        \label{fig:x_v}
    \end{minipage}
    \hfill
    \begin{minipage}[b]{0.46\textwidth}
        \centering
        \includegraphics[width=\linewidth]{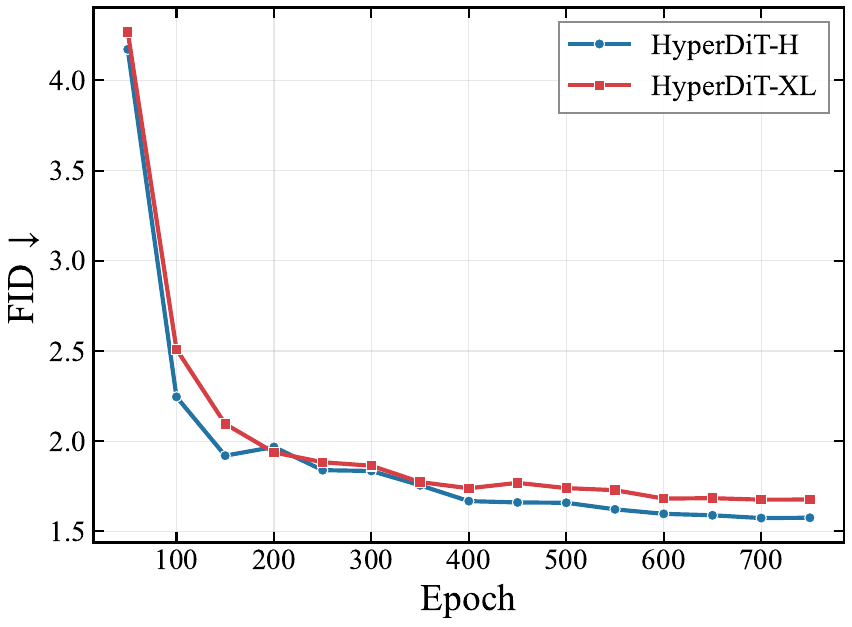}
        \caption{
        FID during training.
        }
        \label{fig:train}
    \end{minipage}
    
\end{figure}

\section{$x$-pred vs. $v$-pred}

We ablate the choice of the training prediction target within our FM formulation. 
Inspired by the findings in JiT \cite{jit}, which advocates for direct data prediction in pixel-space diffusion, we compare velocity prediction ($v$-pred) against clean data prediction ($x$-pred). 
This experiment is conducted on the HyperDiT-XL model trained from scratch, explicitly omitting register tokens and $\mathcal{L}_{REPA}$.

Given the FM trajectory $z_t = t x_0 + (1-t)\epsilon$, the two prediction targets are defined as follows:
\begin{itemize}
    \item $v$-pred: The network directly outputs the velocity $v_\theta(z_t,t)$. The objective is the standard $v$-loss: 
    \begin{equation}
        \mathcal{L}_{v\text{-pred}} = \mathbb{E}_{t,x_0,\epsilon} \left[ \|v_\theta(z_t,t)-(x_0-\epsilon)\|_2^2 \right]
    \end{equation}
    \item $x$-pred: The network is parameterized to directly output the clean image $x_\theta(z_t,t)$, and the optimization is supervised using $v$-loss.
    The predicted velocity is derived as $v_{\theta} = \frac{x_\theta(z_t,t)-z_t}{1-t}$. 
    Thus, the objective becomes:
    \begin{equation}
    \label{eq:x_pred}
        \mathcal{L}_{x\text{-pred}} = \mathbb{E}_{t,x_0,\epsilon} \left[ \left\| \frac{x_\theta(z_t,t)-z_t}{1-t} - (x_0-\epsilon) \right\|_2^2 \right]
    \end{equation}
\end{itemize}

The convergence curves for both configurations are shown in \cref{fig:x_v}. 
We observe that $x$-pred exhibits a noticeably faster convergence rate during the early stages of training (dropping sharply before 100 epochs). 
However, as training progresses, $x$-pred prematurely plateaus. 
In contrast, $v$-pred continues to optimize steadily, eventually surpassing $x$-pred and converging to a lower FID.

\cref{eq:x_pred} introduces a scaling factor of $\frac{1}{(1-t)^2}$ on the MSE of $x_\theta$. 
As $t \to 1$ (approaching the clean data), the gradient aggressively penalizes minute deviations in the absolute pixel prediction. 
While JiT mitigates this in a single-stream model, in HyperDiT, the dense cross-scale attention mechanisms amplify this instability. 
The amplified gradients from the fine-grained flow disrupt the stability of the semantic anchors during joint backpropagation.
$v$-pred, on the other hand, only requires the network to predict the immediate velocity vector $v_t$. 
It is fundamentally easier for the model to infer a local denoising direction from an intermediate noisy anchor than to hallucinate the final clean image.

\section{FID during Training}

To illustrate the learning dynamics of our framework, we plot FID against the training epochs for HyperDiT-XL and HyperDiT-H in \cref{fig:train}. 
Both models exhibit rapid convergence during the initial 200 epochs, followed by a steady and stable decline. 
Notably, the larger variant, HyperDiT-H, consistently outperforms HyperDiT-XL throughout the later stages of training. 
The smooth downward trajectories without severe oscillations explicitly demonstrate the training stability of our proposed pixel-space architecture.

\section{More Visualized Results}
\label{appendix:more_visulized_results}
To further demonstrate the diversity of the dataset, we present additional image samples from LiWi-100k.

\begin{figure}[t]
    \centering
    \includegraphics[width=\textwidth]{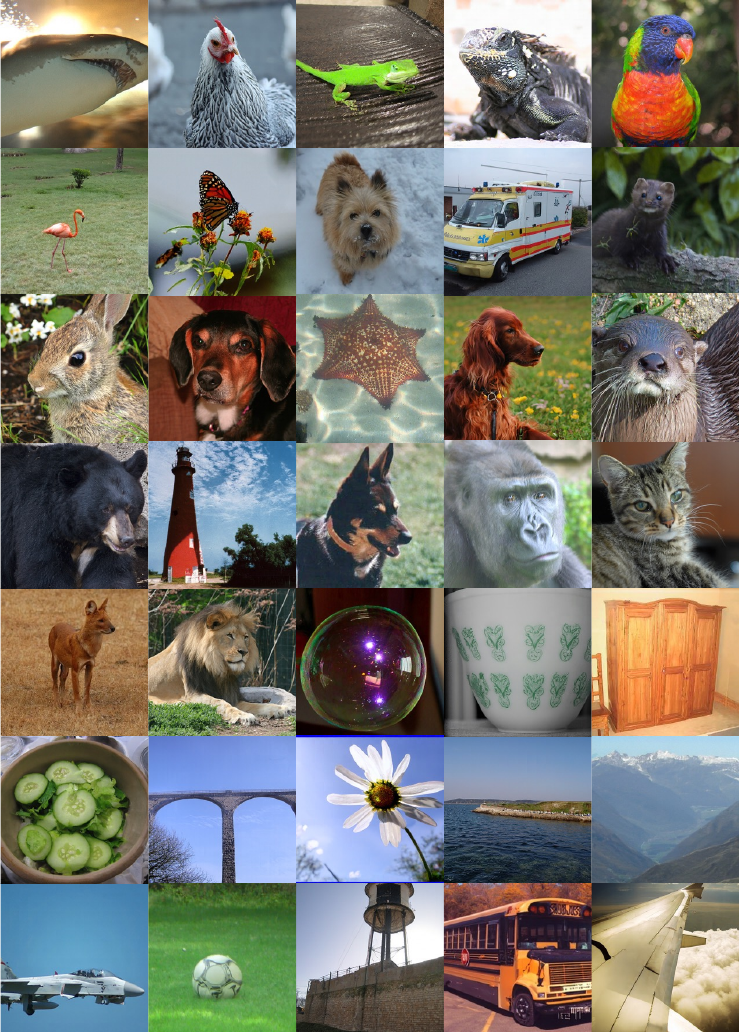}
    \caption{More generated images by HyperDiT-XL at $256\times256$ resolution.}
    \label{fig:more_xl}
\end{figure}

\begin{figure}[t]
    \centering
    \includegraphics[width=\textwidth]{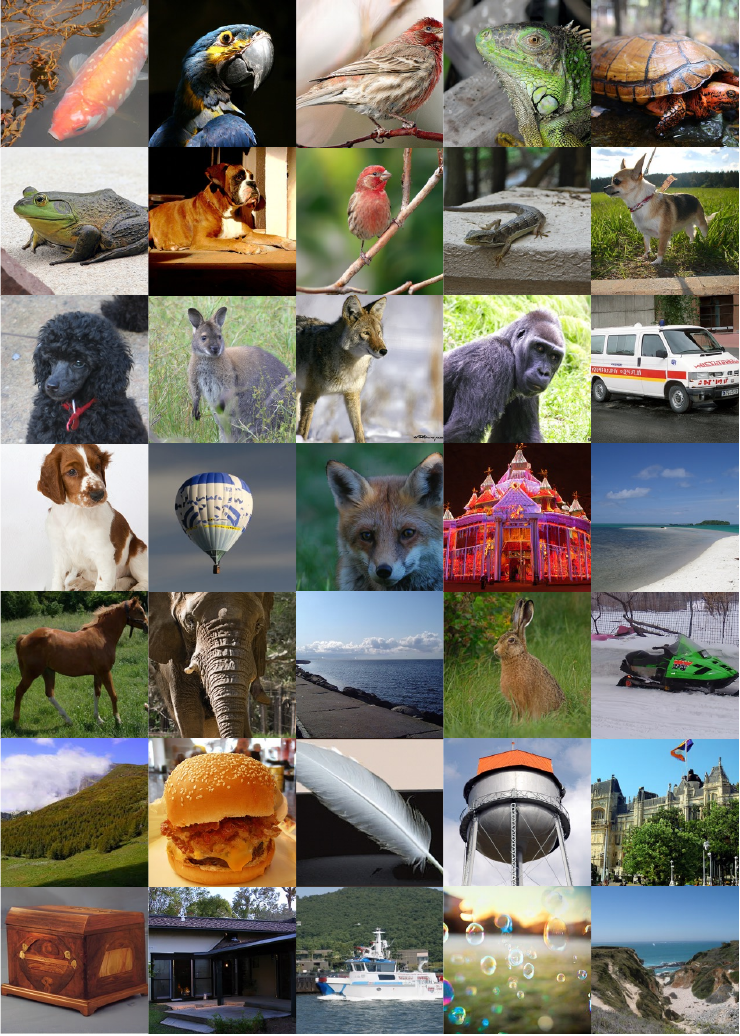}
    \caption{More generated images by HyperDiT-H at $256\times256$ resolution.}
    \label{fig:more_h}
\end{figure}

\cref{fig:more_xl} and \cref{fig:more_h} display the high-fidelity images generated by HyperDiT-XL and HyperDiT-H, respectively, at a resolution of $256 \times 256$. 

\clearpage
\section*{NeurIPS Paper Checklist}


\begin{enumerate}

\item {\bf Claims}
    \item[] Question: Do the main claims made in the abstract and introduction accurately reflect the paper's contributions and scope?
    \item[] Answer: \answerYes{} 
    \item[] Justification: The abstract and introduction clearly state the proposed HyperDiT framework, including Hyper Connectors, SA-RoPE, and register-based dense semantic learning. The main claims are supported by the experimental results in Section 5, including ImageNet 256$\times$256 comparisons and ablation studies.

\item {\bf Limitations}
    \item[] Question: Does the paper discuss the limitations of the work performed by the authors?
    \item[] Answer: \answerYes{} 
    \item[] Justification: Due to the increase in computational complexity with the square of the number of tokens, our method may be difficult to scale to smaller batches, as explained in the result section

\item {\bf Theory assumptions and proofs}
    \item[] Question: For each theoretical result, does the paper provide the full set of assumptions and a complete (and correct) proof?
    \item[] Answer:  \answerNA{} 
    \item[] Justification: Our method does not involve theoretical results

    \item {\bf Experimental result reproducibility}
    \item[] Question: Does the paper fully disclose all the information needed to reproduce the main experimental results of the paper to the extent that it affects the main claims and/or conclusions of the paper (regardless of whether the code and data are provided or not)?
    \item[] Answer: \answerYes{} 
    \item[] Justification: The paper provides the model architecture, training setup, optimizer, learning rate, batch size, number of epochs, hardware type, sampling method, evaluation metrics, and detailed hyperparameters in Section 5 and Appendix A.

\item {\bf Open access to data and code}
    \item[] Question: Does the paper provide open access to the data and code, with sufficient instructions to faithfully reproduce the main experimental results, as described in supplemental material?
    \item[] Answer: \answerYes{} 
    \item[] Justification: We will open-source the code along with documentation for some of the data sources after submission.

\item {\bf Experimental setting/details}
    \item[] Question: Does the paper specify all the training and test details (e.g., data splits, hyperparameters, how they were chosen, type of optimizer) necessary to understand the results?
    \item[] Answer: \answerYes{} 
    \item[] Justification: Section 5.1 and Appendix A specify the ImageNet 256$\times$256 setting, patch sizes, optimizer, learning rate, batch size, number of epochs, GPU type, sampler, sampling steps, CFG scale, and evaluation metrics. Appendix A further reports architecture and sampling hyperparameters for HyperDiT-B, HyperDiT-XL, and HyperDiT-H.

\item {\bf Experiment statistical significance}
    \item[] Question: Does the paper report error bars suitably and correctly defined or other appropriate information about the statistical significance of the experiments?
    \item[] Answer: \answerYes{} 
    \item[] Justification: The FID statistic is computed over 50k samples, and is therefore stable with respect to randomness. For both comparison results and ablation study evaluations, we train multiple trials and report the average results to ensure the statistical significance of the experiments.

\item {\bf Experiments compute resources}
    \item[] Question: For each experiment, does the paper provide sufficient information on the computer resources (type of compute workers, memory, time of execution) needed to reproduce the experiments?
    \item[] Answer: \answerYes{} 
    \item[] Justification: We completed training on an 8×B200 server, with training taking approximately 6 minutes per epoch.
    
\item {\bf Code of ethics}
    \item[] Question: Does the research conducted in the paper conform, in every respect, with the NeurIPS Code of Ethics \url{https://neurips.cc/public/EthicsGuidelines}?
    \item[] Answer: \answerYes{} 
    \item[] Justification: Our paper follows the NeurIPS Code of Ethics.

\item {\bf Broader impacts}
    \item[] Question: Does the paper discuss both potential positive societal impacts and negative societal impacts of the work performed?
    \item[] Answer: \answerYes{} 
    \item[] Justification: Since the work improves high-fidelity image generation, potential positive uses include improved visual synthesis and creative applications, while potential negative uses include misuse for synthetic or deceptive imagery.
    
\item {\bf Safeguards}
    \item[] Question: Does the paper describe safeguards that have been put in place for responsible release of data or models that have a high risk for misuse (e.g., pre-trained language models, image generators, or scraped datasets)?
    \item[] Answer: \answerYes{} 
    \item[] Justification: Our training data will undergo strict filtering, and we will ensure rigorous verification before open-sourcing to check for any obvious issues like harmful visual content.

\item {\bf Licenses for existing assets}
    \item[] Question: Are the creators or original owners of assets (e.g., code, data, models), used in the paper, properly credited and are the license and terms of use explicitly mentioned and properly respected?
    \item[] Answer: \answerYes{} 
    \item[] Justification: We used the DINO model and the JIT codebase for our experiments, and cited them in the corresponding sections of the paper.

\item {\bf New assets}
    \item[] Question: Are new assets introduced in the paper well documented and is the documentation provided alongside the assets?
    \item[] Answer: \answerNA{} 
    \item[] Justification: Our paper does not release new assets.

\item {\bf Crowdsourcing and research with human subjects}
    \item[] Question: For crowdsourcing experiments and research with human subjects, does the paper include the full text of instructions given to participants and screenshots, if applicable, as well as details about compensation (if any)? 
    \item[] Answer: \answerNA{} 
    \item[] Justification: Our paper does not involve crowdsourcing nor research with human subjects.

\item {\bf Institutional review board (IRB) approvals or equivalent for research with human subjects}
    \item[] Question: Does the paper describe potential risks incurred by study participants, whether such risks were disclosed to the subjects, and whether Institutional Review Board (IRB) approvals (or an equivalent approval/review based on the requirements of your country or institution) were obtained?
    \item[] Answer: \answerNA{} 
    \item[] Justification: Our paper does not involve crowdsourcing nor research with human subjects.

\item {\bf Declaration of LLM usage}
    \item[] Question: Does the paper describe the usage of LLMs if it is an important, original, or non-standard component of the core methods in this research? Note that if the LLM is used only for writing, editing, or formatting purposes and does \emph{not} impact the core methodology, scientific rigor, or originality of the research, declaration is not required.
    \item[] Answer: \answerNA{} 
    \item[] Justification: The core method development in this research does not involve LLMs as any important, original, or non-standard components.

\end{enumerate}

\end{document}